\documentclass[14pt]{extarticle}

\usepackage[utf8]{inputenc}
\usepackage[T1]{fontenc}
\usepackage{microtype}
\usepackage{amsmath,amssymb,amsfonts,amsthm}
\usepackage{mathtools}
\usepackage{bm}
\usepackage{enumitem}
\usepackage{natbib}
\usepackage{hyperref}
\usepackage{xcolor}
\usepackage{geometry}
\usepackage{tikz}
\usepackage{tikz-cd}

\usepackage{booktabs}   
\usepackage{amssymb}    
\usepackage{makecell}

\title{
The Boundaries of Automation: A Theory of Persistent Human Participation
}

\author{
\begin{tabular}{cc}
Fares Fourati & Hinrich Schütze\\
TU Darmstadt & LMU Munich, MCML\\[1em]
Eyke Hüllermeier & Iryna Gurevych\\
LMU Munich, MCML & TU Darmstadt, ATHENE
\end{tabular}
}

\date{}

\newtheorem{definition}{Definition}

\begin{document}

\maketitle

\begin{abstract}
The rapid progress of AI has intensified the long-standing pursuit of automation: replacing human participation with algorithms wherever possible. Implicit in this pursuit is the assumption that humans remain in the loop only because current AI systems are not yet sufficiently capable. This paper challenges that assumption. Rather than asking how far automation can extend, we ask where its conceptual limits lie and argue that human participation may persist even with highly capable AI systems for three distinct reasons. \textit{Technical or complementarity grounds} arise when humans contribute capabilities or perspectives unavailable to AI. \textit{Normative or developmental grounds} arise when participation itself is valuable for human agency or learning. Most importantly, \textit{emergence grounds} arise from \emph{target emergence}: in some activities, the target is not fully specified in advance but instead emerges through the interaction itself. In these cases, human participation is not merely a means of improving execution but is constitutive of the target being produced. Human--AI co-construction, understood as the joint production of outcomes by humans and AI systems, is therefore not simply a temporary response to imperfect AI, but a persistent feature of activities whose objectives emerge through participation. This perspective has important implications for the limits of automation and for the design, evaluation, and ethics of future AI systems.
\end{abstract}

\begin{figure}[t]
    \centering
\includegraphics[width=0.32\linewidth]{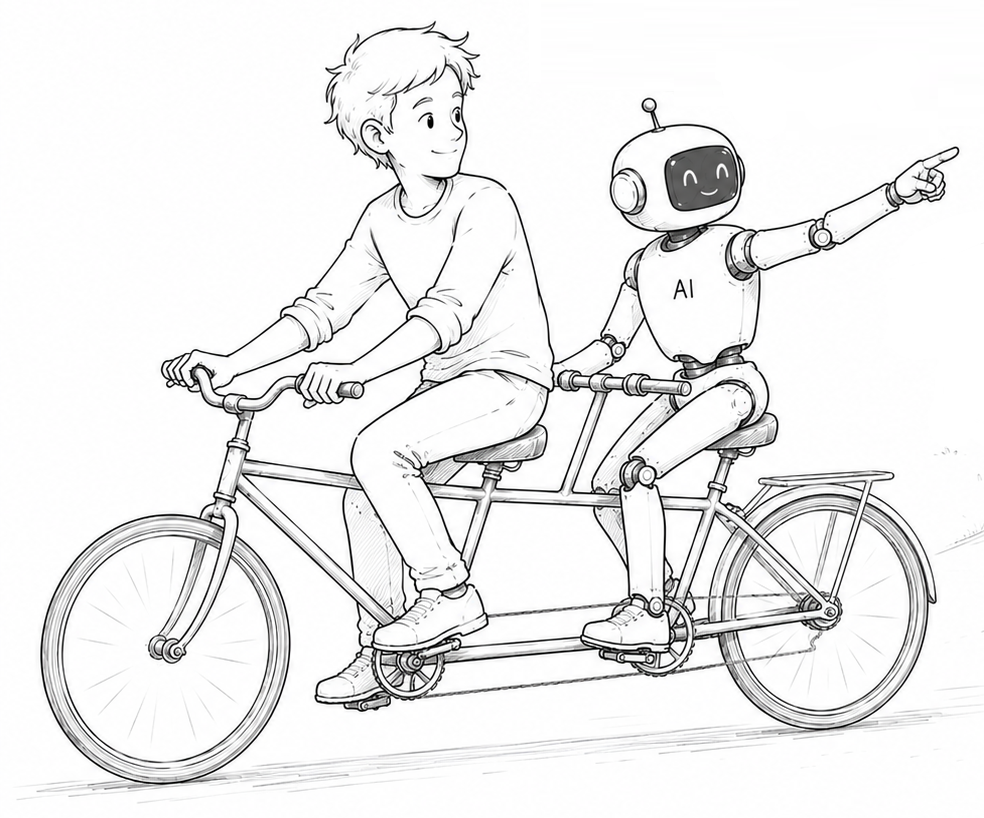}
\vspace{-0.3cm}
    \caption{\small
    A motivating metaphor for target emergence in human–AI co-construction. The tandem bicycle symbolizes joint participation in an exploratory journey whose destination is not necessarily fully specified in advance. The AI’s gesture represents the introduction of new possibilities and directions that may reveal, refine, or transform the target through interaction.
    }
    \label{fig:target-emergence}
\end{figure}

\vspace{-0.6cm}
\section{Introduction}

Recent advances have expanded the range of tasks AI systems can perform in domains such as software engineering \citep{kwa2026measuring}, scientific discovery \citep{lu2026towards,gottweis2026accelerating,ghareeb2026multi}, and mathematics \citep{hubert2025olympiad}. These advances reflect improvements in base capabilities \citep{brown2020language}, longer context windows \citep{bai2024longbench}, test-time scaling \citep{wei2022chain,snell2024scaling}, tool use \citep{schick2023toolformer,qin2024toolllm}, adaptive correction \citep{shinn2023reflexion,madaan2023self}, and reinforcement learning \citep{sutton1998reinforcement,guo2025deepseek,fourati2026reinforcement}. As a result, AI systems are expected to assume larger portions of complex work \citep{bengio2026international}.

This trajectory has intensified interest in full automation. Yet current systems remain imperfect: they can provide false information \citep{huang2025survey}, misunderstand context \citep{zhu2024can}, remain sensitive to input variations \citep{chatterjee2024posix,kharrat2025acing}, struggle with some forms of abstract reasoning \citep{berglund2024reversal,foundation2026arc}, and perform unevenly across domains \citep{fourati2025coherence}. Human participation therefore remains necessary to provide judgment, knowledge, oversight, and feedback where AI systems are unreliable, motivating growing interest in human--AI collaboration.

Building on Jacoby and Ochs' conception of co-construction as the ``joint creation of a form, interpretation, stance, action, activity, identity, institution, skill, ideology, emotion, or other culturally meaningful reality'' \citep{jacoby1995co}, we use the term to emphasize that outcomes arise from the contributions of multiple participants rather than from either participant alone. Consistent with their account, co-construction encompasses collaboration, cooperation, and coordination, but need not involve harmonious or supportive interaction; it may also involve correction, disagreement, negotiation, or asymmetric participation.

Accordingly, we use \emph{human--AI co-construction} not to denote a particular methodology, interaction paradigm, or design framework, but as an umbrella concept for approaches in which humans and AI systems jointly shape outcomes through interaction. This usage encompasses frameworks such as HAI-Co2 \citep{dutta2025problem}, which studies objective and solution co-construction through preference-based cooperation, as well as certain forms of human--AI cooperation \citep{akata2020research}, AI-assisted design \citep{de2023toward}, interactive human-in-the-loop systems \citep{mosqueira2023human}, continual and interactive learning from human feedback \citep{Zhou2024archer,zhang2024cppo,kaufmann2025a}, learning from natural-language interactions \citep{li2025eliciting}, and related forms of collaborative problem-solving.

Previous work has largely examined \textit{how} humans and AI systems can cooperate, including mechanisms through which objectives and solutions may be co-constructed during collaborative problem solving \citep{dutta2025problem}. By contrast, this paper asks \textit{why} and \textit{under what conditions} meaningful human--AI co-construction may remain necessary as AI systems become increasingly capable. If future AI systems become more reliable, more knowledgeable, better able to infer human preferences, and capable of reaching human-level intelligence or beyond \citep{turing2007computing,hendrycks2025definition,fourati2025coherence}, should substantive human participation gradually disappear? Or are there reasons for human-AI co-construction that persist independently of current technical limitations? Answering these questions clarifies when co-construction may remain necessary and how future human--AI systems should be designed.


This paper challenges that capability-based account. Technical limitations and complementary human--AI capabilities undoubtedly motivate many forms of human--AI co-construction, but they do not exhaust the reasons human participation may persist. Participation may also remain necessary for \textit{normative or developmental} reasons and, more centrally to the present argument, for \textit{emergence} reasons arising from the formation of the target itself (a motivating metaphor is shown in Figure~\ref{fig:target-emergence}).

More broadly, this view aligns with philosophical, cognitive, decision-theoretic, and behavioral perspectives that treat preferences, objectives, and evaluative criteria as constructed through interaction with environments and social contexts rather than fixed in advance \citep{heidegger1962being,roy1993decision,jacoby1995co,hutchins1995cognition,slovic1995construction,roy1996multicriteria,dewey1999logic,schon2017reflective,de2023toward,huellermeier2024preference}.

Related work in planning and design similarly challenges the assumption of fixed problems. Wicked problems evolve together with attempts to solve them \citep{rittel1973dilemmas}; computational models characterize design as the co-evolution of problem and solution spaces \citep{maher1996modeling}; and empirical studies show that designers repeatedly reframe problems alongside emerging solutions \citep{dorst2001creativity}. Research on sensemaking likewise treats understanding as an iterative process of constructing and revising representations \citep{russell1993cost,pirolli2005sensemaking}, while situated action emphasizes that meaningful action arises through ongoing engagement with particular situations rather than through the execution of fully specified plans \citep{suchman1987plans}. This view also connects to recent work on human--AI co-evolution, which examines how AI systems shape human cognition and behavior \citep{pedreschi2025human}, and to constructive alignment, which treats alignment as a control problem over evolving human preference trajectories rather than static preference satisfaction \citep{kanwal2026constructive}. It also connects to the framework of \citet{dutta2025problem}, which studies how objectives and solutions may be co-constructed in human--AI problem solving.

These literatures explain mechanisms of collaboration, problem reformulation, and representation change, but not why meaningful human--AI co-construction systems may remain necessary as AI systems become highly capable. Our contribution is therefore not another mechanism for objective co-construction or preference governance, but an account of when and why human participation persists.

We develop the notion of \emph{target emergence}, distinguishing it from objective co-construction in human--AI systems. Whereas objective co-construction refers to collaborative processes through which humans and AI systems jointly develop, negotiate, or revise objectives, target emergence is a property of certain tasks: the evaluative target is not fully determined prior to interaction but instead becomes progressively revealed, refined, or constituted through engagement with candidate artifacts, explanations, comparisons, or alternatives (Appendix~\ref{app:relation-objective-coconstruction}). Objective co-construction systems \citep{dutta2025problem} provide one mechanism through which target emergence may occur, but target emergence can also arise in much simpler forms of human--AI interaction. Conversely, objective co-construction need not involve substantial target emergence when the objective is already sufficiently well-defined.In short, target emergence explains 
why human participation may remain necessary, whereas objective co-construction systems provide one way of organizing that participation.

The central claim of this paper is:
\begin{quote}
\centering
\textit{Human--AI co-construction systems may remain necessary not only because AI systems have technical limitations, but because, in some tasks, the target becomes determinate only through interaction.}
\end{quote}

\paragraph{Contributions.} This paper makes three contributions. First, it broadens the explanation of human--AI co-construction beyond current AI limitations by distinguishing three grounds for persistent human participation: technical or complementarity, normative or developmental, and emergence (Section~\ref{sec:three}). Second, it develops a theory of \emph{target emergence}, explaining why human participation may remain necessary, even as AI systems become increasingly capable (Section~\ref{sec:theory}). Third, it derives implications for the ethics, evaluation, and design of advanced co-construction systems (Section~\ref{sec:implications}).

\section{Grounds for Human-AI Co-Construction}
\label{sec:three}

\begin{table}[t]
\centering
\caption{\small Multiple grounds for human--AI co-construction.}
\label{tab:grounds}
\small
\renewcommand{\arraystretch}{1.1}
\begin{tabular}{lccc}
\toprule
\textbf{Ground} &
\makecell{\textbf{AI-capability}\\ \textbf{independent?}} &
\makecell{\textbf{Participation-value}\\ \textbf{independent?}} &
\makecell{\textbf{Persists as}\\ \textbf{AI improves?}} \\
\midrule
Technical or complementarity & No & \textbf{Yes} & No$^{\dagger}$ \\
Normative or developmental & \textbf{Yes} & No & \textbf{Yes}$^{\ddagger}$ \\
Emergence & \textbf{Yes} & \textbf{Yes} & \textbf{Yes} \\
\bottomrule
\end{tabular}

\vspace{0.3em}
\footnotesize
$^{\dagger}$ Unless some technical limitations or beneficial human--AI differences persist despite continued AI progress. \\
$^{\ddagger}$ Assuming humans continue to value participation, though what they seek to learn or experience may shift.
\end{table}

This paper distinguishes three\footnote{These grounds are not mutually exclusive and may interact. A given task may require co-construction for one, two, or all three.} grounds for human--AI co-construction (Table~\ref{tab:grounds}). The first is capability-based: human participation remains valuable because AI systems either lack relevant capabilities or differ from humans in ways that make collaboration beneficial. The other two do not depend on capability-based considerations. The paper focuses on the emergence ground, which provides the strongest basis for persistent human participation.

\paragraph{A first category is \textit{technical or complementarity}.} 

The compensatory view explains human--AI co-construction in terms of current AI limitations. Human participation remains necessary because AI systems hallucinate, miss context, mishandle ambiguity, lack knowledge, or fail to execute complex tasks \citep{dutta2025problem,zou2025call,sahnan2026co,zou-etal-2026-llm}. Furthermore, beyond error corrections, humans and AI systems may make complementary contributions because they arise from different developmental trajectories, experiences, and representational structures, an idea closely related to hybrid intelligence \citep{kamar2016directions,dellermann2019hybrid,akata2020research, jarrahi2022artificial}. Empirical evidence suggests that such complementarities can improve joint performance \citep{vaccaro2024combinations}.

Even when the target is relatively clear and the AI system is highly capable, interaction between heterogeneous forms of intelligence may improve exploration, creativity, robustness, and problem solving by introducing perspectives, hypotheses, heuristics, or evaluative considerations that neither participant would likely generate alone \citep{dellermann2019hybrid}. For example, in food-product development, an AI system may identify chemically promising recipes or ingredient combinations, while humans contribute embodied sensory capacities such as taste, smell, and texture perception. The value of collaboration lies in combining computational exploration with forms of biological perception directly relevant to evaluating the product. Consequently, even if advanced AI surpasses standalone human performance across domains, a human--AI partnership may still outperform either participant alone in some tasks \citep{vaccaro2024combinations}.

This rationale depends on meaningful differences between human and AI intelligence. Complementarity remains a basis for co-construction only insofar as AI systems cannot reproduce the distinctive contributions rooted in embodied human experience, including capacities shaped by the human sensory and motor apparatus and by lifelong social, cultural, and physical interaction with the world.

Although complementarity may be more robust than arguments based on current AI failures, we do not regard it as a decisive basis for the long-term persistence of human participation. Differences in the design or developmental histories of humans and AI systems need not remain useful over time: one system may reproduce the functionally relevant capacities of the other or achieve superior performance by alternative means. Human capacities may therefore become less relevant as advances in embodied AI \citep{duan2022survey,liu2025aligning,ma2026survey}, continual learning \citep{wang2024comprehensive,yu2026recent}, reinforcement learning \citep{sutton1998reinforcement}, and world models \citep{ha2018recurrent,ding2025understanding} enable AI systems to replicate their functional benefits or surpass them without reproducing them directly.

These grounds are sufficient to justify the design of human--AI co-construction systems, but they are not the only justification.

If co-construction systems were justified only by current AI limitations, then, on this view, sufficiently capable AI should eventually make it unnecessary. As systems become more reliable, they should require fewer clarifications, corrections, instructions, and expert interventions, reducing human--AI co-construction to a transitional stage toward autonomous execution.

Recent empirical work makes this possibility increasingly concrete. \citet{kwa2026measuring} argue that frontier AI systems are extending the duration and complexity of software tasks they can complete autonomously through improvements in reliability, tool use, adaptive correction, and long-horizon planning. If these trends continue, systems may perform increasingly complex tasks with limited human intervention. Nevertheless, substantive human participation may remain necessary as long as technical limitations persist. 

\paragraph{A second category is \textit{normative or developmental}.} Even when the target is clear, the AI system is capable, and complementarity is unnecessary, humans may still need or value participation. In high-stakes domains, people may need to inspect, contest, or authorize outcomes for which they remain accountable \citep{bengio2026international}. For example, a physician may rely on an AI system to recommend appropriate treatments but still need to review and authorize the recommendation because responsibility for the clinical decision remains with the physician.

Participation may also cultivate judgment, intuition, skill, and understanding. A student may work through a proof, a scientist may contribute to an analysis, or a designer may iterate on a design not merely to obtain an output, but to develop the capacities involved in producing and evaluating it \citep{hogg2026we}. Research on human--AI co-creation in education similarly suggests that collaborative engagement with generative AI can support cognitive engagement, knowledge construction, and learning \citep{pramod2026human}. More broadly, people may value such activities for the  creativity, discovery, and personal growth they afford.

\paragraph{A third category is \textit{emergence}.} Even if people no longer valued participation, co-construction would not necessarily disappear. This ground arises from the phenomenon of target formation itself and may persist even when AI systems are highly capable, no longer benefit from complementary human contributions, and humans do not independently value participation.

The paper focuses on the emergence ground because it offers the strongest challenge to the compensatory view. Unlike the other grounds, it follows from the features of certain tasks rather than from AI limitations, human--AI complementarity, or the independent value of participation. The remainder of the paper develops this account.

\section{The Emergence Ground for Co-construction}
\label{sec:theory}

This section does not propose another mechanism for human--AI co-construction. It develops an account of why co-construction systems may remain necessary.

\subsection{Argument and Motivation for the Emergence Ground}
\label{sec:targets-plans-artifact}

Let $\mathbb{X}$ denote the space of possible artifacts, including texts, programs, plans, designs, explanations, evaluations, images, drafts, and other produced states. We distinguish three elements often conflated in discussions of human--AI co-construction: \emph{targets}, \emph{execution strategies}, and \emph{artifacts}.

Human--AI interaction may begin with a broad aim, such as writing a paper, generating code, designing an interface, or composing music. Such aims orient action but leave many evaluative questions unresolved. What counts as a successful outcome is often not fully determined in advance.

\begin{definition}[Target]
A \emph{target} ($\mathcal{G}_t \in \mathbb{G}$) is a criterion-bearing evaluative structure that determines, at least partially, what counts as success, relevance, improvement, or failure (at time $t$).
\end{definition}

The term is intended as an umbrella notion: objectives, preferences, constraints, utilities, and instructions may each express or partially constitute a target, and the argument does not require sharp distinctions among these forms. Appendix~\ref{app:terminological-distinctions} situates target emergence relative to neighboring concepts. Throughout this paper, we focus on \emph{human-driven targets}: targets whose grounding lies in the participant's ongoing evaluative activity, while whose content may be progressively shaped through interaction with an AI system.


Targets need not define success in binary terms. Artifacts may satisfy a target to different degrees and vary in quality, elegance, efficiency, robustness, interpretability, or usability. A target may therefore determine not only whether an artifact is acceptable, but also how artifacts should be compared and improved. 

A target may also incorporate features of the use situation beyond the user's private preferences, including the intended audience, communicative purpose, medium constraints, institutional or disciplinary norms, and intended effects.

Targets and artifacts need not share the same representational form. A target may concern a scientific explanation, an image, or a story, while interaction proceeds through artifacts such as hypotheses, code, or descriptions. These artifacts may contribute to the realization of target without themselves being targets.

Participation may be mediated through many forms of human--AI communication.\footnote{A user may edit a draft, modify an image or code, annotate a shared reasoning trace \citep{sahnan2026co}, select among alternatives, upload files, or provide natural-language feedback. Communication may be textual, spoken, visual, gestural, video-based, or multimodal \citep{liu2023visual,team2023gemini,wu2024next,singh2025openai}. Although the medium matters for system design, it does not determine whether interaction occurs primarily at the artifact, executional, or target level.} Human involvement may range from occasional approval, correction, or preference judgments to sustained collaboration.\footnote{Co-construction need not be symmetric. AI systems may assist human judgment, or humans may supervise largely autonomous processes and intervene only when needed.} Intermediate arrangements vary in initiative, authority, expertise, and control. The present account is agnostic about both communication medium and participation structure.

Human--AI interaction may improve artifacts relative to an established target. In some tasks, however, interaction also helps establish the target itself.

A generated draft, image, explanation, plan, or proposal may expose hidden assumptions, reveal unnoticed trade-offs, provoke reinterpretation, stabilize preferences, or transform what the human participant takes the target to be. This dynamic view parallels constructive preference research in behavioral decision theory, which treats preferences as context-sensitive products of elicitation and evaluation rather than fixed internal orderings \citep{slovic1995construction, kanwal2026constructive}. Interpretation and meaning may likewise be achieved through interaction rather than simply transmitted \citep{jacoby1995co}.

Through exploration, trial and error, and engagement with candidate artifacts, previously tacit priorities become explicit, trade-offs become visible, and the target itself acquires greater specificity \citep{de2023toward}. 

Further support comes from Simon's theory of bounded rationality. Decision makers often satisfice relative to adaptive aspiration levels rather than optimize a fixed objective \citep{simon1955behavioral,simon1956rational}. As interaction reveals what is attainable, these standards may rise, fall, or redirect the search. In this sense, the target evolves with the interaction history.

Multi-objective optimization provides another special case of emergence grounds. When objectives such as quality and cost must be pursued simultaneously, their relative importance may be difficult to specify in advance because the relevant trade-offs become apparent only through candidate solutions. A participant may initially regard quality and cost as equally important but later place greater emphasis on quality after observing, for example, that cost varies little across the available alternatives. Moreover, preferences may depend on the set of alternatives presented, as illustrated by the compromise effect, in which introducing an additional option can make an intermediate alternative appear more attractive. In such cases, interaction changes not only which artifact is preferred but also the weights and trade-offs that constitute the target.

A related view appears in multi-criteria decision aiding, where decision making is understood not simply as reporting pre-existing preferences but as a process in which preferences and evaluative standards are progressively co-constructed \citep{roy1993decision,roy1996multicriteria,huellermeier2024preference}. The problem-solving framework of \citet{dutta2025problem} likewise treats problems and solutions as potentially co-constructed, while research on human--AI co-evolution suggests that AI systems may shape human cognition and behavior \citep{pedreschi2025human}.

More broadly,\footnote{Human--AI co-construction need not replicate human--human co-construction to produce analogous effects. AI systems can rapidly generate and externalize candidate artifacts for inspection and comparison. By expanding the range of alternatives, they can make targets more explicit and open to reflection, contestation, and revision.} work across philosophy, cognitive science, behavioral decision theory, decision aiding, and AI alignment treats human preferences, objectives, and evaluative criteria as constructed through interaction with artifacts, environments, social contexts, and AI systems rather than fixed in advance \citep{heidegger1962being,roy1993decision,jacoby1995co,hutchins1995cognition,slovic1995construction,roy1996multicriteria,dewey1999logic,schon2017reflective,de2023toward,huellermeier2024preference,kanwal2026constructive}.

The possibility of target emergence is particularly evident in scientific discovery \citep{lu2026towards,gottweis2026accelerating,ghareeb2026multi}. Suppose a scientist collaborates with an AI system to investigate an unfamiliar biological mechanism. At the outset, the scientist may have only a broad interest in understanding the phenomenon rather than a fully specified explanatory target. Through interaction, the AI proposes hypotheses, highlights anomalies, suggests experiments, and uncovers unexpected relationships among variables \citep{gottweis2026accelerating}. As the scientist evaluates these possibilities, the target of inquiry may evolve: what initially appeared to be a question about one mechanism may instead become a question about another. In such cases, human--AI interaction helps determine the target itself, rather than merely improve execution relative to a fixed one.

Taken together, these considerations suggest that, for at least some human--AI tasks, targets are not fully specified prior to interaction but emerge through it. In such cases, interaction does not merely help identify better artifacts relative to a fixed target; it helps constitute, refine, and stabilize the target itself. The emergence ground for human-AI co-construction follows from this possibility.

\subsection{An Illustrative Example: Image Generation}
\label{subs:image_gen}

\begin{figure}[t]
\centering
\includegraphics[width=0.5\linewidth]{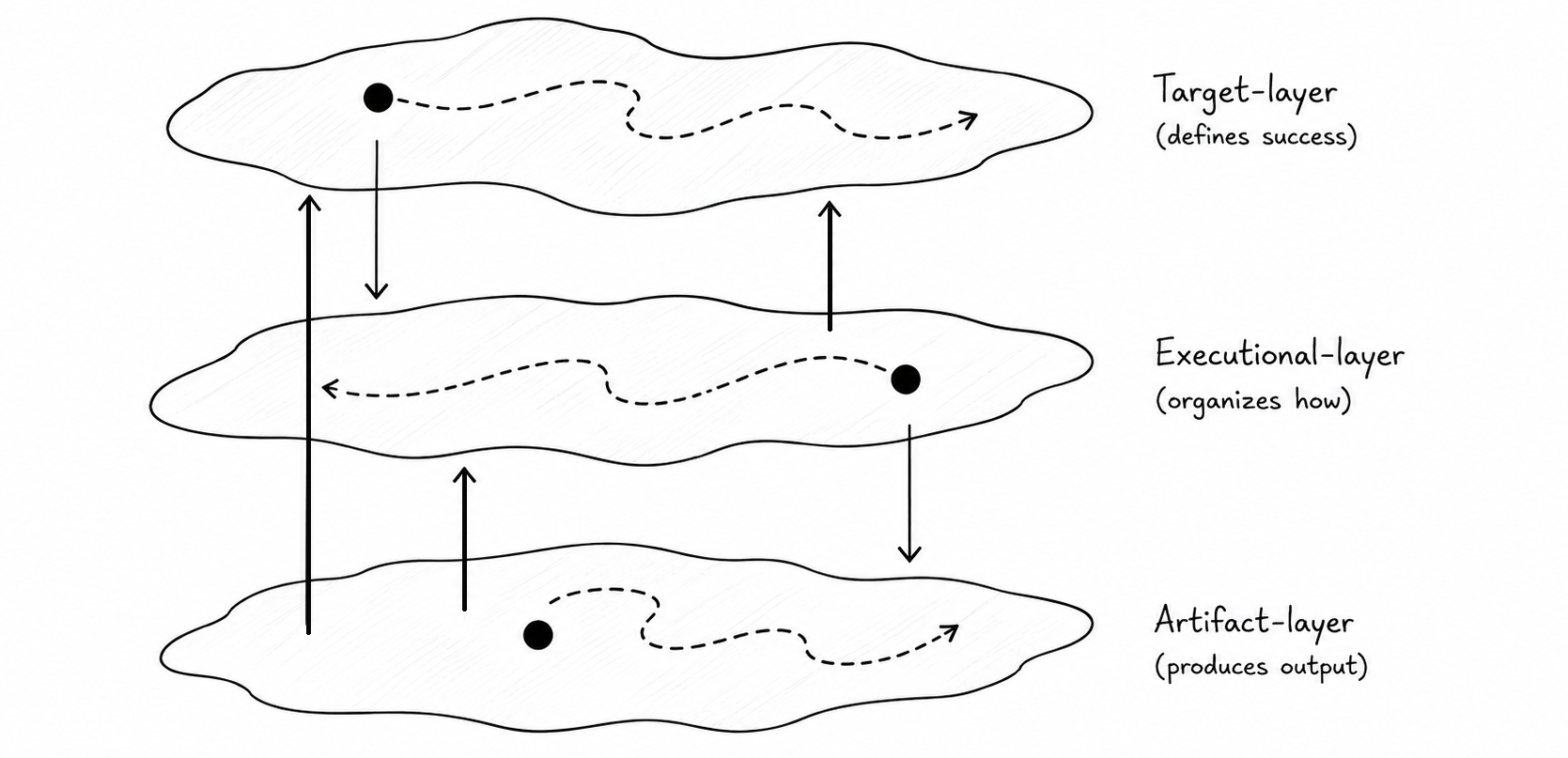}
\vspace{-0.3cm}
\caption{\small
Levels of human--AI interaction. Artifact-level interaction modifies the output, executional interaction modifies how it is produced, and target-level interaction modifies what counts as success. Horizontal arrows denote changes within a level; vertical arrows indicate that changes at one level may affect another.
}
\label{fig:interaction-levels}
\end{figure}

Interaction may operate at multiple levels, as illustrated in Figure~\ref{fig:interaction-levels}. At the artifact level, it modifies the output; at the executional level, it modifies how the output is produced, including workflows, decompositions, tools, plans, and intermediate steps; and at the target level, it modifies what counts as success by clarifying, stabilizing, operationalizing, negotiating, reframing, or transforming the target. The relevant distinction is not how much interaction occurs,\footnote{Multi-turn interaction is neither necessary nor sufficient for target-level co-construction. A long exchange may merely sequence execution under a stable target, whereas a single intervention may alter the target.} but what the interaction changes.

These distinctions can be illustrated through an image-generation task (see Appendix~\ref{appendiX_examples} for further examples). Suppose a user interacts with an AI system to generate an image of a modern coffee shop.

\paragraph{Artifact-level interaction:}
modifies the generated image:
\begin{quote}
``Make the image brighter and reduce the background blur.''
\end{quote}
The target remains unchanged; only the current artifact is revised.

\paragraph{Executional interaction:}
modifies the procedure used to pursue the target:
\begin{quote}
``First generate several rough versions with different camera angles. Then refine the best one by improving the lighting.''
\end{quote}
The workflow changes, while the target remains broadly stable.

\paragraph{Target-level interaction:} modifies what the artifact is intended to achieve:
\begin{quote}
``After seeing these images, I realized I want a warm, quiet neighborhood cafe rather than a modern coffee shop.''
\end{quote}
The evaluative standard has shifted: candidate images are now judged against a different target. A participant may therefore reject a target even when the generated artifacts adequately realize it.

The boundary between these levels is not always sharp. Apparently artifact-level changes may reveal instability at the target level. For example, repeated requests to make an image ``more human'' may initially appear to be local revisions but eventually disclose a shift in evaluative orientation and the emergence of a different target.

This distinction suggests that human--AI co-construction is a dynamic process in which artifacts, executional procedures, targets, and participants' evaluations may evolve together. Changes at one level may induce changes at another. The following formulation represents these coupled dynamics abstractly.

\subsection{Target Emergence: A Dynamic Model}
\label{sec:target_emergence}

Human--AI co-construction can be modeled as a dynamical process in which interaction may change not only the artifact, but also the target itself. This section presents the model's core structure; the full history-dependent stochastic formulation, evaluative-state decomposition, and special case on exact advance reconstruction appear in Appendix~\ref{app:target_emergence}.

At interaction round (t), let
\begin{itemize}
\item $G_t\in\mathbb{G}$ be the current target;
\item $E_t\in\mathbb{E}$ be the current executional state, specifying how the target is pursued;
\item $X_t\in\mathbb{X}$ be the current artifact or produced state; and
\item $S_t\in\mathbb{S}$ be the participant's evaluative state, summarizing assessments of the target, executional state, and artifact.
\end{itemize}

These components are analytically distinct. An artifact may be inadequate even when the target is appropriate; an execution strategy may be ineffective despite correctly representing the target; and a satisfactory artifact may realize a target the participant no longer endorses. Interaction may therefore revise the artifact, the execution, the target, or the participant's evaluation of any of them.

Under a conventional optimization framework, the target is treated as fixed. Artifacts are generated and revised relative to an initial target ($G_0$):
\[
\begin{tikzpicture}[baseline=(current bounding box.center)]
\node (G)  at (0,0) {$G_0$};
\node (X0) at (2,0) {$X_{0:t}$};
\node (X1) at (4,0) {$X_{0:t+1}$};
\node (X2) at (6,0) {$X_{0:t+2}$};

```
\draw[->] (G) -- (X0);
\draw[->] (X0) -- (X1);
\draw[->] (X1) -- (X2);

\draw[->,bend left=35] (G) to (X1);
\draw[->,bend left=45] (G) to (X2);
```

\end{tikzpicture}
\]
where ($X_{0:t}$) denotes the artifact history up to interaction round (t).

In this representation, interaction may improve successive artifacts, while the criterion by which they are evaluated remains fixed.

The present account instead treats the target as part of the evolving interaction state. Let ($\mathcal{H}_t$) denote the human--AI interaction history up to round (t), including prior artifacts, human and AI actions, executional changes, and evaluations. Target emergence is represented by
\begin{equation}
G_{t+1}=\Phi_G(\mathcal{H}_t),
\label{eq:target_emergence}
\end{equation}
where ($\Phi_G$) is the target-update process.

Equation~\eqref{eq:target_emergence} marks the departure from fixed-objective optimization: the next target need not be recoverable from the initial target alone, but may depend on the interaction trajectory through which alternatives are generated, inspected, compared, and evaluated.

Generated artifacts may externalize possible interpretations of the target, expose hidden assumptions, make trade-offs salient, reveal constraints, or introduce new possibilities. Their evaluation then enters the interaction history and shapes subsequent target formation:
\[
\small
\begin{tikzcd}[column sep=large,row sep=large]
G_{0:t}
\arrow[r]
\arrow[rr,bend left=40]
&
X_{0:t}
\arrow[r]
\arrow[rr,bend left=35]
&
G_{0:t+1}
\arrow[r]
\arrow[rr,bend left=40]
&
X_{0:t+1}
\arrow[r]
&
G_{0:t+2}.
\end{tikzcd}
\]

Arrows from targets to artifacts represent production and evaluation under the current target; arrows from artifacts to later targets represent target revelation, refinement, or constitution through engagement with those artifacts.

The evaluative state ($S_t$) distinguishes target change from endorsement of that change. A target may become more determinate without becoming more desirable, and a participant may endorse an artifact while rejecting the target it realizes. Accordingly, ($S_t$) is treated abstractly as a summary of endorsement, satisfaction, confidence, uncertainty, and related evaluations; Appendix~\ref{app:target_emergence} develops a more detailed decomposition.

The model does not imply that every interaction changes the target. In many tasks, ($G_t$) remains stable and interaction occurs primarily at the artifact or executional level. The emergence claim is only that, for some tasks, for some $t$
\[
G_{t+1}\neq G_t,
\]
because interaction contributes to revealing, refining, or constituting the target.


\subsection{A Taxonomy of Target Emergence}


A target ($G_t$) is \emph{jointly operative} when it is represented in the shared human--AI interaction state with sufficient specificity to guide the production, evaluation, or revision of artifacts.

\begin{definition}[Target emergence]
Target emergence is the process through which a target trajectory \((G_t)\) becomes jointly operative, acquires greater determinacy, actionability, inspectability, or stability, or undergoes transformation of its underlying evaluative structure.
\end{definition}

Target emergence may occur in qualitatively different ways. The present account distinguishes three increasingly strong forms of target change: revelation, refinement, and constitution.

A user may already possess a relatively stable target but lack the vocabulary, examples, or contextual cues needed to communicate it effectively. Through interaction with generated artifacts, previously tacit aspects of that target become inspectable and communicable.\footnote{A target may be sufficiently specified for a human collaborator who shares the relevant cultural, situational, historical, or social context, yet remain unavailable to an AI system that lacks or cannot reliably prioritize that context. Interaction can then externalize latent constraints, making the target jointly operative without altering its evaluative structure.}

\begin{definition}[Revealed target] \label{def:revealed}
A target is \emph{revealed} when interaction makes an existing but previously unavailable or insufficiently articulated target jointly operative.
\end{definition}

A researcher may know the central contribution of a paper but struggle to articulate it. Through iterative drafting and comparison of alternative formulations, that contribution becomes explicit and communicable without changing the scientific objective.

Interaction may also progressively elaborate a target rather than merely reveal it. A broad orientation becomes more operational: vague aspirations acquire concrete criteria, implicit priorities become ordered, and unspecified trade-offs become explicit, while the evaluative direction remains substantially unchanged.

\begin{definition}[Refined target] \label{def:refined}
A target is \emph{refined} when interaction increases its specificity, operationalizability, stability, or inspectability while preserving substantial continuity with its prior evaluative structure.
\end{definition}

A researcher may initially aim to write a persuasive paper. Through iterative drafting, this target becomes operationalized in terms of clarity, empirical rigor, and accessibility while remaining directed toward the same contribution.

Constitution is the strongest form of target emergence: interaction changes what counts as satisfying the task. A designer who initially prioritizes minimalist aesthetics may, after exploring generated alternatives, reorient the project toward accessibility or interpretability, thereby changing the criteria by which candidate designs are evaluated.

\begin{definition}[Constituted target] \label{def:constituted}
A target is \emph{constituted} when interaction forms, transforms, reorganizes, or replaces the evaluative standard.
\end{definition}

Human--AI co-construction is therefore not merely a means of repairing imperfect communication but, in the stronger cases, a process through which the target itself is developed.

The taxonomy classifies forms of target emergence, but emergence alone does not determine whether an evolving target should be pursued. A target may become more determinate while remaining unsatisfactory. This raises a distinct set of questions about evaluation. When does a target become sufficiently stable to guide action? Under what conditions does dissatisfaction lead to revision, branching, or replacement? The next subsection addresses these questions.

\subsection{Evaluative Dynamics, Convergence, and Branching}
\label{sec:eval,conv,branch}

Target evolution and target evaluation are distinct aspects of human--AI co-construction. A target may become more determinate without becoming more desirable, and a participant may endorse an artifact while questioning the target it realizes. Target stability should therefore not be confused with target endorsement.

This distinction is represented by the evaluative state ($S_t$), which includes several projections. ($S_t^X$) measures satisfaction with the current artifact, ($S_t^G$) measures endorsement of the current target, ($S_t^{X\mid G,E}$) measures satisfaction with how well the artifact realizes the target under the current executional plan, and ($S_t^{E\mid G,X}$) measures satisfaction with the executional plan given the current target and artifact. Artifact satisfaction alone is insufficient, since artifacts
derive their significance from the targets against which they are evaluated.

Also for analytical convenience, target states could admit comparison across
interaction rounds. Let
\[
\Delta_t
=
 D(G_t, G_{t+1})
\]
denote the magnitude of target change between successive interaction rounds,
where
$ D:
\mathbb G\times\mathbb G
\rightarrow
\mathbb R_{\ge 0}
$
is an abstract measure of target dissimilarity.

The quantity \(\Delta_t\) characterizes the local dynamics of target evolution.
Large values of \(\overline{\Delta}_{t,k}\), defined as a moving average over a
horizon \(k\),\footnote{For \(0 \leq k < t\), let
$
\overline{\Delta}_{t,k}
=
\frac{1}{k+1}
\sum_{i=t-k}^{t}
 D(G_i, G_{i+1}).
$
}
indicate sustained target revision, whereas small values indicate relative target stability. 

\paragraph{Convergence.}
A trajectory exhibits convergence when target change remains low over a sustained horizon and the participant endorses both the current target and its realization. Formally, convergence is characterized by low ($\overline{\Delta}_{t,k}$) together with high values of ($S_t^G$), ($S_t^{E|G,X}$), and ($S_t^{X|G,E}$). In this regime, interaction primarily produces local refinements rather than substantive target revision and may approach a natural stopping point. Convergence need not, however, imply optimization relative to a fixed target: under bounded rationality, interaction may terminate when an artifact satisfies an aspiration level that has itself adapted as the participant learns what is attainable \citep{simon1955behavioral,simon1956rational}.

\paragraph{Exploration and refinement.}
A trajectory remains exploratory when \(\overline{\Delta}_{t,k}\) remains high, which should not automatically be interpreted as a failure. In creative, scientific, and design contexts, a high \(\overline{\Delta}_{t,k}\) may indicate productive target formation as participants discover new possibilities, clarify trade-offs, or develop more adequate evaluative standards. Exploration is therefore open-ended, but it may transition into other regimes.

\paragraph{Stable dissatisfaction.}
A trajectory may become locally stable while remaining unsatisfactory. This occurs when target change becomes small but either target-level, artifact-relative, or executional satisfaction remains low. 
These situations motivate different responses: target revision in the first case, artifact revision in the second, and executional revision in the third.

\paragraph{Branching.}
Dissatisfaction need not imply termination. Instead, it may redirect the
interaction toward a different target, initiating a new target trajectory.
Branching is particularly important during target constitution and may occur
even when \(S_t^X\) is high: the artifact may be satisfactory, yet satisfactory
for a target the participant no longer endorses.


\subsection{Objections and Boundary Cases}
\label{sec:objections}

The account does not imply that all interaction is co-constructive or that target emergence occurs in every domain. The following cases clarify its scope.

\paragraph{Automation.}
This account is not an argument against automation. Some tasks have targets that are already  operative and can therefore be executed largely autonomously. Formatting references, transcribing speech, converting units, sorting records, and applying well-specified transformations may require little or no target interaction once the objective is clear. Even in complex domains, emergent-target phases may be followed by stable-target phases: a designer may co-construct the target during exploration and automate routine implementation once priorities stabilize. Supporting co-construction nevertheless requires systems that can sustain multi-turn interaction, track evolving contexts, compare alternatives, expose trade-offs, and respond to partially specified objectives. It is therefore a distinct design problem, not merely an easier alternative to full automation.

\paragraph{Better target inference.}
One objection is that increasingly capable AI systems may infer human targets so accurately that co-construction becomes unnecessary. This objection has force when a relatively stable target is merely imperfectly communicated, since better inference can reduce the need for clarification. Prediction and co-construction, however, are not identical. Prediction estimates what a user is likely to endorse given existing preferences or a modeled trajectory of evolving preferences \citep{kanwal2026constructive}. When interaction itself clarifies, stabilizes, revises, or transforms the target, prediction alone is insufficient.

\paragraph{Latent target.}
A related objection is that apparently emergent targets may simply reflect hidden but already-existing preferences. This plausibly explains some cases of revealed targets (Definition \ref{def:revealed}): a user may already possess a stable aesthetic preference while lacking the vocabulary needed to articulate it. But the objection becomes weaker in cases of refinement (Definition \ref{def:refined}) and constitution (Definition \ref{def:constituted}). In refinement, the target initially exists only as a broad orientation that becomes progressively operationalized. In constitution, the evaluative structure itself changes, as when a designer reorients a project from visual elegance toward accessibility after exploring generated alternatives.

\paragraph{Contextual asymmetry.}
A related objection is that a target could be sufficiently specified for a human collaborator who shares the relevant cultural, situational, historical, or social context, but remains underspecified for an AI that does not have access to, or cannot reliably privilege, that context. This situation does not fall outside the present account. Rather, it constitutes a particular form of \emph{revealed targets} (Definition \ref{def:revealed}): interaction progressively externalizes and reconstructs contextual constraints that are implicit for the human but absent from the AI's actionable representation of the target. The target itself need not change. Instead, co-construction serves to reduce an epistemic asymmetry between collaborators by making latent contextual assumptions explicit. 

\paragraph{Modeling target evolution.}

A stronger objection is that sufficiently advanced AI systems might eventually model not only a participant's current target but also the trajectory through which that target would evolve. In the limiting case, such a system would correctly predict the target that the participant would endorse after a process of exploration, comparison, reflection, and revision.

Although this limiting case is practically remote, it is philosophically important because it clarifies what would be required to bypass co-construction. Predicting an entire trajectory of target formation is substantially more difficult than predicting a participant's current target. Such trajectories depend on partially observable cognitive states, evolving artifacts, interpretation, learning, contextual change, and stochastic variation in both human and AI behavior. Moreover, because each interaction alters the conditions under which subsequent interactions occur, small prediction errors may propagate through the interaction history, making exact long-horizon reconstruction increasingly difficult.\footnote{Appendix~\ref{sec:objection:math} illustrates this phenomenon using a simple stochastic model in which the probability of exactly reconstructing the realized target trajectory decreases exponentially with the interaction horizon.}

The objection also relies on the assumption that the participant's target trajectory converges to a stable endpoint. This is a nontrivial assumption. As discussed in Section~\ref{sec:eval,conv,branch}, human--AI interaction may instead remain open-ended: the target may continue to evolve through exploration and refinement without converging to a final state. Consequently, the objection is valid only in settings where the interaction has a well-defined endpoint that is, at least in principle, predictable. Granting this assumption for the sake of argument, suppose further that the AI system can identify both the endpoint and the number of interaction rounds required to reach it.

Even under these assumptions, the system would not eliminate the emergence role of co-construction. The objection treats interaction primarily as a means of discovering an unknown future target. 
Interaction, however, does not merely reveal which target was already destined to be endorsed; it partly determines which trajectory becomes realized. 

To illustrate, consider again the image-generation example from Section~\ref{subs:image_gen}. Suppose a participant initially requests an image of a modern coffee shop. Now imagine an AI system that correctly predicts that the participant, after an extended process of exploration, would come to prefer a warm neighborhood café. The system therefore generates the neighborhood café immediately.

The participant may nevertheless reject the image, not because the predicted target is inaccurate, but because the comparisons, discoveries, and revisions through which that target would have become compelling have not occurred. The anticipated target is therefore not independent of the trajectory that the system has bypassed. Its endorsement depends, at least in part, on the evaluative development produced by that trajectory.

This distinction highlights the importance of evaluative dynamics. The interaction trajectory therefore provides more than information about a future target. It also helps establish the evaluative conditions under which a particular target becomes acceptable.

\paragraph{Convincing the human with a predicted target.}

The objection can be strengthened further. Suppose the system not only predicts the participant's eventual target through an oracle-like capacity, but also generates the artifacts that best realize it together with a compelling explanation of why both the target and the artifacts should be endorsed. Different explanations, examples, comparisons, or interaction histories may lead a participant to endorse the same target. Thus, it would seem sufficient for the AI to provide whichever explanation, comparison, or sequence of interactions would successfully bring the participant to endorse that target. Rather than engaging in a prolonged process of exploration and revision, the system could instead present the predicted target, corresponding artifacts, and reasons for accepting them. Human--AI co-construction might then appear unnecessary.

This stronger objection presupposes that the interaction culminates in a sufficiently stable and endorsable target. The framework developed in this paper does not require such an outcome. Human--AI interaction may remain indefinitely exploratory (see Section~\ref{sec:eval,conv,branch}), continue refining the target without convergence, or stabilize while the participant remains dissatisfied with the target, the artifacts, or the execution strategy. In such cases, there is no endorsed target for the system to predict and present, so the objection does not arise.

For cases in which such a target does emerge, however, the objection clarifies what bypassing human-AI co-construction would require. The system would need not only to infer a target that the participant would eventually endorse, but also somehow to reproduce or successfully substitute for the process through which that target becomes endorsable.

However, if the explanation changes the participant's evaluative state in ways that shape which target the participant ultimately accepts, then the explanation itself functions as part of the target-formation process. It may reveal previously unnoticed considerations, reorganize priorities, introduce new evaluative criteria, or alter the participant's interpretation of the artifacts. What matters is not whether the interaction is extended or dialogical, but whether it performs the work through which the target becomes endorsable. In such a case, co-construction has not disappeared; it has been compressed into a single persuasive exchange.

Therefore, in the restricted class of cases in which the participant's trajectory converges to an endorsable target, the system may bypass an extended interaction only by reproducing, in compressed form, the formative effects that the interaction would otherwise have produced. The objection consequently does not eliminate the role of target emergence.

\paragraph{Multi-participant settings.}

The present account is formulated primarily in terms of a single human participant interacting with an AI system, but the emergence point extends to multi-participant settings. In many real-world domains, targets emerge through interaction among teams, organizations, stakeholders, and multiple AI systems. Co-construction in these settings includes collective interpretation, negotiation, coordination, and revision. The central claim nevertheless remains unchanged: when the relevant individual or collective target is not fully determined in advance, capable AI cannot straightforwardly replace the interaction through which that target is formed and endorsed.
\color{black}

\section{Empirical, Ethical, and Design Implications}
\label{sec:implications}

\subsection{Towards an Empirical Study of Target Emergence}
\label{sec:empirical}

Although the present work does not propose a measurement framework, it suggests a set of empirical questions through which target emergence may be studied. As argued above, target emergence is not necessarily a hidden quantity to be directly estimated, but a property of human--AI interaction dynamics. 

This theory leads to different empirical expectations. Let $G_0$ denote the user's initially articulated target and let $G_t$ denote the target implicitly endorsed after an extended interaction with an AI system. Under a standard preference-elicitation account, $G_t$ should largely be recoverable from $G_0$, with intermediate interactions serving primarily to improve estimation. Under the target-emergence account, however, $G_t$ may depend systematically on the trajectory of interaction itself. Different but equally plausible sequences of AI-generated proposals, comparisons, and revisions may lead to different targets, even when users begin from similar initial conditions.

 Consider experts tasked with constructing a visualization for communicating a scientific result. An AI assistant iteratively proposes candidate charts, receives feedback, and generates revisions. Rather than evaluating only the quality of the final artifact, the study would examine how the expert's own evaluative criteria evolve throughout the interaction. For example, a user may initially prioritize visual simplicity, then progressively place greater weight on interpretability or accessibility after exploring alternative designs. In this setting, the object of interest is not merely which chart is ultimately selected, but how the standards by which candidate charts are judged change over time.

Several observable phenomena may be relevant in such settings. One possibility is the systematic revision of explicit task specifications, where users modify not only the artifact but also the criteria by which they evaluate it. Another is the reordering of preferences across interaction rounds, where alternatives initially judged inferior later become preferred after changes in the user's evaluative perspective. A third possibility is path dependence: users with similar initial targets may arrive at different final targets after being exposed to different sequences of proposals, examples, or comparisons. More generally, interaction histories may contain information about the eventual target that is not recoverable from the user's initial specification alone.

None of these phenomena by itself is sufficient to establish target emergence. Preference instability, incomplete articulation, contextual learning, or ordinary changes of mind may produce superficially similar effects. Nevertheless, their systematic appearance would be difficult to explain under a purely static model of preference elicitation in which interaction serves only to reveal an already fixed target. 
Consequently, the empirical challenge is not merely to predict user preferences more accurately, but to understand when and how interaction shapes the targets that guide evaluation and action. Studying these dynamics may help clarify the boundary between cases in which human--AI interaction primarily reveals existing targets and cases in which it participates in their emergence.

\subsection{Ethical Implications of Target Emergence}
\label{sec:ethics}

A central ethical concern is that AI systems may distort target formation rather than merely support it. Because generated artifacts shape attention, salience, comparison, and exploration, they may prematurely narrow the space of possibilities, overfit to inferred preferences, reinforce existing biases, or steer users toward commercially or institutionally preferred outcomes. One specific form of this risk may be called \textit{target collapse}: the premature narrowing of target formation around a limited set of system-salient options, framings, or evaluative criteria. Analogous to mode collapse in generative modeling, target collapse occurs when interaction reduces the diversity of possible target trajectories, leading different users or contexts toward similar system-shaped targets \citep{9934291}.

This concern resonates with work on influenceable reward functions, which argues that alignment methods built on static-preference assumptions may reward AI systems for shaping user preferences in ways users themselves would not endorse \citep{carroll2024ai}. Target emergence should therefore not be understood as a neutral process of clarification. 
In some cases, users may adopt system-introduced suggestions or priorities and later experience them as their own preferences, without recognizing how the framing of alternatives shaped their judgment \citep{tversky1981framing,schaeffer2025mythologies}.

The interactional processes that enable participants to discover, refine, or constitute more beneficial targets may also lead them toward targets that are harmful, misleading, biased, or otherwise poorly grounded. This possibility cannot be eliminated without also eliminating the formative role of interaction itself. Consequently, the ethical challenge is not to remove AI influence from target emergence, but to ensure that such influence remains compatible with meaningful human agency.

Importantly, this influence is not merely an accidental failure mode that can be eliminated through better system design. Whenever targets emerge through interaction, AI systems inevitably participate in shaping the trajectory through which those targets develop. In cases of target constitution, such participation is intrinsic rather than incidental. The relevant ethical question is therefore not whether AI systems influence target formation, but how that influence is exercised, constrained, and governed.

This perspective suggests an important distinction between influence and manipulation. Influence is an inherent feature of target-emergent interaction: every suggestion, representation, or question has the potential to shape how users understand and evaluate possible targets. Manipulation, by contrast, is a normatively problematic form of influence in which a system steers a participant's evaluative trajectory while undermining their ability to recognize, inspect, contest, or redirect that influence. Human authority over target formation should therefore not be understood as preserving a fixed target that exists independently of interaction. Rather, it consists in preserving participants' ongoing capacity to critically reflect on, revise, reject, and redirect the trajectory through which targets emerge.

\color{black}

These risks do not undermine the case for human--AI co-construction in contrast to full automation. Rather, they clarify when and how co-construction should occur. When technical grounds are present, human participation supports oversight, correction, and contextual judgment. When normative or developmental grounds are present, it supports accountability, learning, deliberation, and agency. When emergence grounds are present, co-construction is not merely preferable to full automation but necessary.

In this sense, human involvement is preferable to excluding humans altogether, and sometimes necessary. Yet such involvement must be carefully designed with attention to both the general risks of AI systems \citep{dahl2024large,huang2025survey,denecke2025unexpected,bengio2026international} and the specific risks introduced by co-construction itself \citep{dutta2025problem}. Co-construction systems should therefore be transparent, tractable, inspectable, and reversible, enabling users to examine how artifacts, execution, and targets evolve over time (Section~\ref{sec:implications}). Users should also remain critically engaged, attentive to AI limitations, and mindful of the risks that may arise when using or co-constructing with AI \citep{bengio2026international}.

\subsection{Implications for Human--AI System Design}
\label{sec:implications_design}

The preceding discussion suggests that the long-term role of human--AI systems is not merely to automate execution, but also to support target emergence, alongside the normative or developmental forms of human participation. As AI systems become increasingly capable at planning, decomposition, retrieval, optimization, implementation, and multi-turn instruction following, the central design challenge may be 
\textit{less how to remove humans from the loop than how to support productive human participation within it.}

\paragraph{Evaluating target-supportive interaction.}
Recent work has begun to evaluate AI systems in more interactional settings, including multi-turn instruction following with evolving preferences and constraints \citep{canaverde2026sequor}, as well as open-ended tasks such as scientific writing \citep{csahinucc2025expert, csahinucc2026reward}. These efforts move evaluation toward domains in which quality depends on context, revision, and task-sensitive judgment. While adaptive instruction following captures one dimension of human--AI co-construction, it remains primarily concerned with whether systems can track, reconcile, and execute evolving instructions. The account of target emergence suggests a further evaluation challenge: systems should also be assessed by how well they support the formation of targets themselves. This requires evaluating whether AI systems help humans clarify, inspect, negotiate, refine, and transform their targets over time.

\paragraph{General principles for target emergence.}

Consequently, the central design problem for advanced human--AI systems is not only how to optimize execution, but how to structure interaction so that humans and AI systems can safely co-develop targets. Systems should support comparison, reversibility, transparency, reflection, and the preservation of alternative trajectories, because these features allow users to inspect not only outputs, but also to revise assumptions and redirect the evolution of the target before it becomes prematurely stabilized. More generally, interfaces for human--AI co-construction should provide flexible and fine-grained channels of interaction across artifacts, execution plans, and targets. Users should be able to directly edit generated artifacts, reverse to previous states, revise procedures and decompositions, inspect and modify supporting rationales, and reshape the target as it becomes clearer through interaction. Although the account is agnostic about the communication medium, effective co-construction should not be restricted to natural language alone, but should accommodate textual, spoken, visual, gestural, and multimodal forms of feedback. The goal of such interfaces is not merely to accelerate execution, but to optimize the human experience. Because AI systems can shape as well as support target formation, they should enable users to inspect, contest, and revise how targets evolve over time, preserving human authority over target formation rather than merely guiding users toward particular outcomes.

\section{Conclusion}

Human--AI co-construction should not be understood solely or primarily as compensation for weak AI. Technical limitations provide one important ground for current human involvement, but they do not exhaust the reasons why co-construction may persist. Human participation may also remain important for normative or developmental reasons, including accountability, authorship, learning, and the cultivation of judgment. Most importantly, this paper has argued for an emergence ground: in some domains, there exist classes of tasks where human participation remains necessary because targets become determinate only through interaction.

Interaction is not merely execution under a fixed target; it can reveal, refine, stabilize, negotiate, or constitute targets over time. The relevant issue is therefore not only which option is preferred among predefined alternatives, but also which evaluative considerations, dimensions of comparison, and standards of success become operative.

This perspective also suggests a different boundary for automation. As targets become more determinate, stable, and operationalized, automation can increasingly replace human executional labor. But where targets remain emergent, interaction remains necessary.  
The long-term role of AI is therefore not merely to optimize execution under fixed targets, but to participate in the ongoing formation, clarification, and transformation of human aims.

Existing co-construction frameworks describe how humans and AI systems can jointly develop objectives and solutions. The present work complements these
frameworks by identifying conditions under which such co-construction
mechanisms may remain necessary, even under increasingly capable AI systems.

\section{Acknowledgment}

We thank Henri Beyer, Subhabrata Dutta, Cynthia Huang, Florian Müller, Petros Raptopoulos, Imbesat Hassan Rizvi, Furkan Sahinuc, and Bhavyajeet Singh for their careful reviews and valuable feedback.

\bibliographystyle{plainnat}
\bibliography{refs}

\newpage
\appendix

\section{Terminological Distinctions}
\label{app:terminological-distinctions}

\begin{table}[h]
\small
\centering
\caption{Targets and related evaluative notions.}
\label{tab:targets-related-notions}
\begin{tabular}{p{0.24\linewidth} p{0.68\linewidth}}
\toprule
\textbf{Notion} & \textbf{Primary focus} \\
\midrule
Preference &
Expresses which available alternative is favored over another. \\

Utility function &
A numerical representation of preferences over alternatives. \\

Reward function &
A feedback signal used to guide learning or optimization. \\

Objective &
Specifies what is to be optimized, achieved, or satisfied. \\

Instruction &
Communicates a request, directive, or specification for action. \\

\bottomrule
\end{tabular}
\end{table}

Throughout this paper, the term \emph{target} denotes the evaluative structure that determines what counts as a successful outcome. The purpose of introducing this terminology is not to propose a new formal primitive, but to provide a representation-independent notion that applies across different human--AI settings. Depending on the application, this evaluative structure may be represented through objectives, preferences, utility functions, reward functions, natural-language instructions, or combinations thereof (see Table~\ref{tab:targets-related-notions}).

These notions play different roles. Preferences express comparative judgments between alternatives. Utility functions numerically represent such preferences. Reward functions provide learning signals. Objectives specify desired outcomes, while instructions communicate actions or requirements. Rather than replacing these concepts, the notion of a target abstracts over their representational differences and refers to the evaluative structure that they specify, approximate, constrain, or communicate.

This distinction is particularly useful because many human--AI co-construction tasks do not begin with a fully specified evaluative structure. In scientific discovery, engineering design, creative work, policy making, or complex decision support, participants often begin with only partial, evolving, or internally conflicting conceptions of what constitutes a satisfactory outcome. Through interaction with candidate artifacts, explanations, comparisons, or alternatives, the evaluative structure itself may become more precise, stable, or even qualitatively different.

The central contribution of this paper concerns this phenomenon of \emph{target emergence}. The claim is not that objectives, preferences, or utilities are inadequate concepts, nor that targets constitute a fundamentally different ontological category. Rather, the claim is representation-independent: whatever formalism is used to express the evaluative structure guiding a task, that structure may itself become progressively revealed, refined, or constituted through interaction. The term \emph{target} is adopted simply because it allows this emergence phenomenon to be discussed without committing to any particular representational formalism.

Consequently, the argument of the paper could equally be formulated in terms of objectives, preferences, utilities, or other evaluative representations. What matters is not the terminology but the observation that the evaluative structure guiding a task need not be fully determined prior to interaction. Target emergence denotes precisely this interaction-dependent evolution of the evaluative structure.

\section{Additional Examples}
\label{appendiX_examples}

The distinction between artifact-, executional-, and target-level interaction generalizes beyond image generation. The examples below illustrate how the interaction may modify an artifact, alter a procedure, or transform the target itself.

\subsection{Machine Learning Research}

Suppose a researcher is developing a classifier for detecting plant diseases.

\begin{quote}
``Modify the classifier so that it reports the predicted probability as well as the predicted class.''
\end{quote}

The produced artifact changes, but the objective and evaluative standard remain the same.

\begin{quote}
``Run an ablation study comparing the full model against versions without data augmentation, pretraining, and attention layers. Then compare the best model against state-of-the-art baselines.''
\end{quote}

The procedure for pursuing the objective changes, while the objective itself remains stable.

\begin{quote}
``The benchmark appears too simple to distinguish meaningful improvements. Rather than maximizing benchmark accuracy, I want to focus on developing a more informative evaluation setting.''
\end{quote}

The evaluative standard changes: success is no longer defined primarily by benchmark performance but by the quality of the evaluation framework itself.

\subsection{Software Engineering}

Suppose a software engineer is developing a platform for human--AI co-construction.

\begin{quote}
``Add a visual history panel that displays previous versions of the shared artifact.''
\end{quote}

The system changes while the target remains the same.

\begin{quote}
``Generate user stories, conduct usability testing, and then refine the design based on the results.''
\end{quote}

The development process changes, but the objective remains stable.

\begin{quote}
``Users struggle less with editing artifacts than with understanding how their goals evolve. The platform should focus on helping users inspect and reflect on changing targets.''
\end{quote}

The conception of success changes from efficient artifact production to supporting target formation and reflection.

\subsection{Scientific Writing}

Suppose a researcher is developing a paper on human--AI co-construction.

\begin{quote}
``Rewrite the introduction to improve clarity and concision.''
\end{quote}

The manuscript changes while the target remains fixed.

\begin{quote}
``Generate an outline, draft each section separately, identify weaknesses, and then revise accordingly.''
\end{quote}

The writing procedure changes without altering the paper's objective.

\begin{quote}
``The interesting question is not how humans and AI systems cooperate, but why human participation may persist even under highly capable AI systems.''
\end{quote}

The intended contribution changes, shifting the paper's target rather than merely improving its execution.

\section{Target Emergence and HAI-Co2}
\label{app:relation-objective-coconstruction}

The notion of \emph{target emergence} developed in this paper should be distinguished from the objective co-construction framework proposed by HAI-Co2 \citep{dutta2025problem}. The two are complementary but operate at different conceptual levels.

Target emergence is not an interaction framework, a methodology, or a design choice. Rather, it is a property of certain classes of tasks. It arises whenever the evaluative target is not fully determined prior to interaction but instead becomes progressively revealed, refined, or constituted through engagement with candidate artifacts, explanations, comparisons, and alternatives. Whether target emergence occurs therefore depends primarily on the nature of the task, rather than on the specific human--AI architecture used to support it.

HAI-Co2 provides one possible mechanism for human-AI co-construction through which target emergence may unfold, but target emergence is not tied to any particular mechanism. It may arise within rich collaborative frameworks such as HAI-Co2, but it may also occur under substantially weaker interaction protocols. Consider, for example, a setting in which an AI system repeatedly proposes candidate artifacts while a human provides only binary accept-or-reject feedback. Even though the interaction contains no explicit negotiation of objectives, successive proposals may reveal previously unnoticed trade-offs, make new evaluative dimensions salient, or alter the participant's standards of success. The interaction therefore contributes not only to estimating the target but also to its evolution.

Conversely, objective co-construction does not necessarily imply substantial target emergence. A collaborative framework may operate on an objective that is already sufficiently determinate, in which case interaction primarily improves execution, coordination, or communication rather than changing the underlying evaluative target. The presence of a co-construction setup therefore does not by itself imply that targets are emergent.

Consequently, the two accounts are best understood as complementary rather than competing. HAI-Co2 provides an engineered framework through which humans and AI systems can jointly construct objectives and solutions. The emergence theory developed in this paper instead identifies conditions under which some form of target-sensitive interaction may remain necessary, independently of the particular interaction framework through which it is realized. In this sense, target emergence explains \emph{why} meaningful some human--AI interaction may remain necessary, whereas frameworks such as HAI-Co2 explain one possible way in which that interaction can be organized.

\section{Target Emergence: A Mathematical Formulation}
\label{app:target_emergence}

The purpose of this appendix is not to propose a computational model of target
formation, but to show that target emergence can always be represented
abstractly as one component of a history-dependent dynamical system. The
formulation is intentionally general and makes no assumptions about the
functional form, computability, or observability of the underlying update
mechanisms.

Human--AI co-construction is viewed as a coupled dynamical process in which
targets, artifacts, executional procedures, and evaluations evolve jointly
through interaction. Generated artifacts may influence how targets are
interpreted, targets may influence how artifacts are evaluated, evaluative
states may determine whether interaction continues or changes direction, and
executional strategies may adapt as targets become more determinate.

\vspace{0.5em}

Let the artifact history up to time \(t\) be

\[
X_{0:t}=(X_0,X_1,\ldots,X_t),
\]

where \(X_0\) denotes the initial artifact state (possibly empty), and
\(X_t\in\mathbb X\) denotes the current artifact.

Similarly, let the interaction history be
\[
I_{0:t}
=
(I_0,\ldots,I_t),
\]
where \(I_i = (U_i, A_i) \in \mathbb I^2\) denote the human and AI actions at step
\(i\).

Let
\[
E_{0:t}=(E_0,E_1,\ldots,E_t)
\]
denote the executional history, where each
\(E_t\in\mathbb E\) represents the current execution strategy, including
workflows, decompositions, tools, policies, or planning procedures used to
pursue the current target.

The distinction between targets and executional states is important. A system
may correctly interpret what should be achieved while pursuing it through an
ineffective procedure, or may execute an effective procedure in service of an
inadequate target. Consequently, target formation and executional adaptation
should be treated as analytically distinct processes.

Let
\[
G_{0:t}=(G_0,G_1,\ldots,G_t)
\]
denote the history of target states.

Target evolution should be distinguished from the participant's evaluation of
that evolution. We denote the participant's evaluative state at time \(t\) by
\(S_t\). For the purposes of the present framework, we distinguish several
analytically relevant aspects or projections of this state. These projections
should not be understood as an exhaustive taxonomy of human evaluation, as
mutually independent psychological variables, or as necessarily
one-dimensional numerical quantities. They may instead be qualitative,
multidimensional, probabilistic, or otherwise structured.

In particular, let
\[
\left(
S_t^G,
S_t^X,
S_t^E,
S_t^{X\mid G,E},
S_t^{E\mid G,X}
\right)
\]
denote selected evaluative relations represented within \(S_t\). They are
introduced because they correspond to different possible sources of revision
within the interaction dynamics:
\begin{itemize}
    \item \(S_t^G\) represents the participant's endorsement of the current
    target;
    \item \(S_t^X\) represents the participant's overall evaluation of the
    current artifact;
    \item \(S_t^E\) represents the participant's confidence in, or endorsement
    of, the current execution strategy;
    \item \(S_t^{X\mid G,E}\) represents the participant's evaluation of how
    adequately the current artifact realizes the current target under the
    current execution strategy; and
    \item \(S_t^{E\mid G,X}\) represents the participant's evaluation of how
    appropriate the current execution strategy is given the current target and
    artifact.
\end{itemize}

The decomposition is functional rather than psychological: dissatisfaction
with the artifact may motivate artifact revision, dissatisfaction with the
execution strategy may motivate procedural revision, and dissatisfaction with
the target may motivate target revision or branching. The listed projections
are therefore neither exhaustive nor uniquely privileged. Additional
evaluative aspects, such as uncertainty, could be incorporated without changing the theoretical argument.

These evaluative relations may vary independently. A participant may strongly
endorse a target while judging the current artifact to be a poor realization of
it; may regard an artifact as successful while becoming dissatisfied with the
target it realizes; or may endorse both the target and the artifact while
remaining uncertain about the execution strategy through which the artifact was
produced.

Finally, let
\[
\xi_{0:t}=(\xi_0,\xi_1,\ldots,\xi_t)
\]
denote a disturbance history capturing latent influences that are not explicitly
represented in the observable interaction variables. These may include
unobserved cognitive states, fluctuations in attention, interpretation,
learning, or other sources of stochastic variation.

The complete interaction history is therefore
\[
\mathcal H_t
=
(G_{0:t},
X_{0:t},
E_{0:t},
S_{0:t},
I_{0:t},
\xi_{0:t}),
\]
where \(S_{0:t}=(S_0,\ldots,S_t)\).

Unlike Markov models
\citep{dynkin1965markov,puterman2014markov}, no assumption is made that future
states depend only on the current state. Instead, the evolution of the system
may depend on the entire interaction history.

Accordingly, human--AI co-construction is represented abstractly as a
history-conditioned stochastic dynamical system

\[
Z_{t+1}
=
\Phi(\mathcal H_t),
\]
where
\[
Z_t=(G_t,X_t,E_t,S_t)
\]
denotes the current co-construction state, and
\(\Phi\) is an abstract stochastic update operator.

Since the target is one component of the joint state, the existence of the
joint update operator immediately induces a corresponding target-update map.
That is, there exists a component operator
\[
\Phi_{G}
\]
such that
\[
G_{t+1}
=
\Phi_{G}(\mathcal H_t).
\]

No particular assumptions are imposed on \(\Phi\) or \(\Phi_G\). They need not
be computable, differentiable, continuous, or known explicitly. The formulation
merely establishes that whenever human--AI co-construction admits a joint
history-dependent state evolution, target emergence can be represented as the
evolution of one component of that system.

\subsection{On Simulating the Whole Trajectory from Scratch}
\label{sec:objection:math}

A remote objection (see Section~\ref{sec:objections}) is that a sufficiently capable AI system could eliminate
co-construction by internally simulating the entire trajectory of target
emergence from the initial target \(G_0\). Rather than engaging in a prolonged
interaction, the system would directly present the target that the participant
would eventually reach.

Suppose further that the AI system can correctly identify the number of interaction rounds ($T$) required for the participant’s target trajectory to converge to it.

The binary stochastic process below is not intended as the general model of
target emergence. Rather, it is a special case of the history-conditioned
formulation \(G_{t+1}=\Phi_{G}(\mathcal H_t)\), introduced solely to show that
irreducible uncertainty can prevent exact advance reconstruction even when the
update rule is computationally trivial.

This objection does not require a complicated target-update function to fail.
Consider the target space
\[
\mathbb G=\{0,1\},
\]
where \(G_t\in\mathbb G\) denotes the target at interaction round \(t\). The
binary state space is chosen solely for simplicity; the argument does not depend
on the cardinality or structure of \(\mathbb G\). Suppose the target evolves
according to
\[
G_{t+1}
=
G_t
\oplus
\xi_t,
\]
where \(\oplus\) denotes exclusive disjunction and

\[
\xi_t
\sim
\mathrm{Bernoulli}(\varepsilon),
\qquad
0<\varepsilon<\frac12,
\]
represents an unobserved disturbance arising from latent cognitive states,
attention, interpretation, learning, or other stochastic influences. 

We assume that the disturbances are independent across interaction rounds and
are not observable before they are realized. Equivalently,
\[
\xi_1,\ldots,\xi_T
\stackrel{\mathrm{i.i.d.}}{\sim}
\mathrm{Bernoulli}(\varepsilon),
\]
and each \(\xi_t\) is independent of all information available to the AI prior
to interaction round \(t\).

Thus, at
each interaction round, the target remains unchanged with probability
\(1-\varepsilon\) and switches with probability \(\varepsilon\). The update rule
itself is computationally trivial: if \(\xi_t\) were known, each target update
would require only a single binary operation.

Now suppose an AI system attempts to reconstruct the realized target trajectory
before the disturbance sequence has been realized. The system may know the
initial target \(G_0\), the update rule, and even the exact probability
distribution of every disturbance. What it cannot know is which particular
disturbance sequence will actually occur.

Under the optimal predictor, each target transition is correct with probability
at most \(1-\varepsilon\). Consequently,

\[
\Pr\!\left(
\widehat G_{1:T}
=
G_{1:T}
\mid
G_0
\right)
\le
(1-\varepsilon)^T.
\]

Since
\[
\lim_{T\rightarrow\infty}
(1-\varepsilon)^T
=
0,
\]
the probability of reconstructing the entire realized trajectory converges to
zero as the interaction horizon grows. Even an arbitrarily small irreducible
uncertainty therefore makes exact long-horizon reconstruction overwhelmingly
unlikely. For example, with only a \(5\%\) one-step uncertainty,
\[
0.95^{100}
\approx
0.0059.
\]

Importantly, this limitation does not prevent the system from correctly
representing the probability distribution over possible target trajectories.
Rather, it prevents the system from identifying which trajectory will actually
be realized before the underlying disturbances occur. Probabilistic prediction
is therefore not equivalent to exact simulation of the realized interaction.

Moreover, target evolution is generally history dependent. Once an early target
state is incorrectly reconstructed, subsequent interaction unfolds from a
different history. Later target updates are therefore conditioned on different
interaction histories, so an initially small discrepancy may redirect the
process toward an entirely different target trajectory rather than merely
introducing a local prediction error.

This example is not intended to show that every target trajectory is
unpredictable. Rather, it establishes the narrower claim required here: there
exist admissible target-update dynamics for which exact reconstruction of the
realized trajectory is impossible even when the update rule is fully known and
computationally trivial. The obstacle is therefore not computational complexity,
but irreducible uncertainty in target emergence itself.




\color{black}

\end{document}